\documentclass[letterpaper, 10 pt, conference]{ieeeconf}  %
\IEEEoverridecommandlockouts                              %

\usepackage{amsmath,amssymb}
\usepackage{flushend}
\usepackage{lipsum}
\usepackage[utf8]{inputenc}
\usepackage{graphicx}
\usepackage{color}
\usepackage{latexsym}
\usepackage{bmpsize}
\usepackage{xurl}
\usepackage{comment}
\usepackage[nobreak]{cite}
\usepackage[subrefformat=parens]{subcaption}
\usepackage[hyperfootnotes=false]{hyperref}
\usepackage{cleveref} %
\usepackage{here}
\usepackage{listings} %
\usepackage[affil-it]{authblk} %
\usepackage{mathrsfs} %
\usepackage{algcompatible}
\usepackage{algorithm}
\usepackage{siunitx}

\crefname{figure}{Fig.}{Fig.}%
\crefname{table}{Table}{Table}%
\crefname{section}{Section}{Section}%
\newcommand{\todo}[1]{}
\renewcommand{\todo}[1]{{\color{red} {\bf TODO:#1}}}
\newcommand{\circletext}[1]{\raise0.2ex\hbox{\textcircled{\scriptsize{#1}}}}

\def\Underline{\setbox0\hbox\bgroup\let\\\endUnderline}
\def\endUnderline{\vphantom{y}\egroup\smash{\underline{\box0}}\\}

\definecolor{myComment}{rgb}{0.0, 0.6, 0.0}       %
\definecolor{myKeyWord}{cmyk}{1.0, 0.0, 0.0, 0.3} %
\definecolor{myString}{cmyk}{0.0, 1.0, 0.0, 0.0}  %

\lstdefinestyle{customText}{
    backgroundcolor  = {\color{white}},               %
    basicstyle       = {\footnotesize},               %
    breaklines       = {true},                        %
    commentstyle     = {\itshape  \color{myComment}}, %
    keywordstyle     = {\bfseries \color{myKeyWord}}, %
    lineskip         = {-0.5ex},                      %
    showstringspaces = {false},                       %
    sensitive        = {true},                        %
    stepnumber       = {1},                           %
    stringstyle      = {\ttfamily \color{myString}},  %
    tabsize          = {2},                           %
    xleftmargin      = {2zw},                         %
    xrightmargin     = {2zw}                          %
}
\begin{document}

\title{\bf{\LARGE{{Time-Optimal Path Tracking with ISO Safety Guarantees}}}}
\author[1,2]{Shohei Fujii\thanks{E-mail: \href{mailto:SHOHEI001@e.ntu.edu.sg}{SHOHEI001@e.ntu.edu.sg}; Corresponding author} }
\author[1,3]{Quang-Cuong Pham}
\affil[1]{School of Mechanical and Aerospace Engineering, Nanyang Technological University, Singapore}
\affil[2]{DENSO CORP., Japan}
\affil[3]{Eureka Robotics, Singapore}
\maketitle

\begin{abstract}
  One way of ensuring operator's safety during human-robot collaboration is through Speed and Separation Monitoring (SSM), as defined in ISO standard ISO/TS 15066.
  In general, it is impossible to avoid all human-robot collisions: consider for instance the case when the robot does not move at all, a human operator can still collide with it by hitting it of her own voluntary motion.
  In the SSM framework, it is possible however to minimize harm by requiring this: \emph{if} a collision ever occurs, then the robot must be in a \emph{stationary state} (all links have zero velocity) at the time instant of the collision.
  In this paper, we propose a time-optimal control policy based on Time-Optimal Path Parameterization (TOPP) to guarantee such a behavior.
  Specifically, we show that: for any robot motion that is strictly faster than the motion recommended by our policy, there exists a human motion that results in a collision with the robot in a non-stationary state.
  Correlatively, we show, in simulation, that our policy is strictly less conservative than state-of-the-art safe robot control methods.
  Additionally, we propose a parallelization method to reduce the computation time of our pre-computation phase (down to about 0.5 sec, practically), which enables the whole pipeline (including the pre-computation) to be executed at runtime, nearly in real-time.
  Finally, we demonstrate the application of our method in a scenario: time-optimal, safe control of a 6-dof industrial robot. %

\end{abstract}

\section{Introduction} \label{sec:intro}
Productivity is crucial in robotic automation, whereas the safety of human workers who work with collaborative robots side-by-side must be ensured.
To guarantee the safety of human workers who work with collaborative robots side-by-side, ISO standards define four technical specifications~\cite{ISOTS15066, villaniSurveyHumanRobot2018}.
Among them, in terms of performing automation tasks, they regulate the robot's speed when human workers approach in Category 3 - Speed and separation monitoring (SSM), and the robot's force and torque to minimize injuries caused by collision in Category 4 - Power force limiting (PFL).
Our focus in this paper is to both guarantee the safety in the context of SSM and maximize the speed/productivity simultaneously.
Specifically, we consider a time-optimal trajectory/path tracking problem where a robot moves along a given path as \emph{fast} as possible, and the robot \emph{safely} stops at the time instant of the collision with humans not to harm them.

One of the state-of-the-art methods is proposed by Zanchettin et al.~\cite{zanchettinSafetyHumanrobotCollaborative2016}, which computes the maximum velocity on every control cycle based on current states such as positions and velocities of a robot and obstacles during the path tracking.
This method does not require the entire tracking path to be given before the execution.
Therefore, advantageously, the path is switchable at runtime to, for example, generate an evasive or reactive motion as long as it satisfies continuity.
However, it is also a disadvantage because the method does not exploit the entire path information and fully-leverage the capability of the robot, which makes the system conservative and less productive.

\begin{figure}[t]
  \begin{center}
  \includegraphics[width=1.0\linewidth,trim={1cm 2cm 0.8cm 4cm},clip]{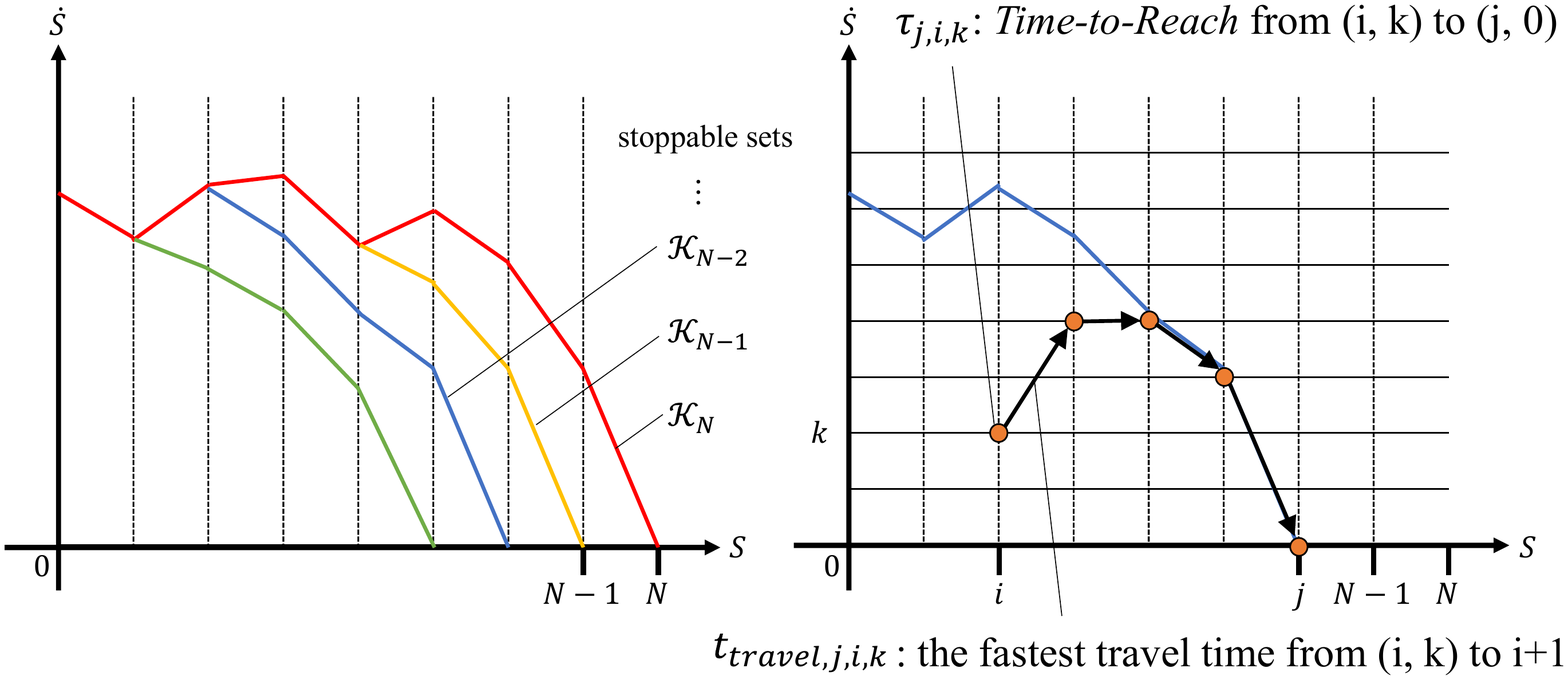}
  \vspace{-10mm}
  \subcaption{}
  \label{fig:method_precomputation}
  \end{center}

  \vspace{-5mm}
  \begin{center}
  \includegraphics[width=0.95\linewidth,trim={7.9cm 2.8cm 1cm 3cm},clip]{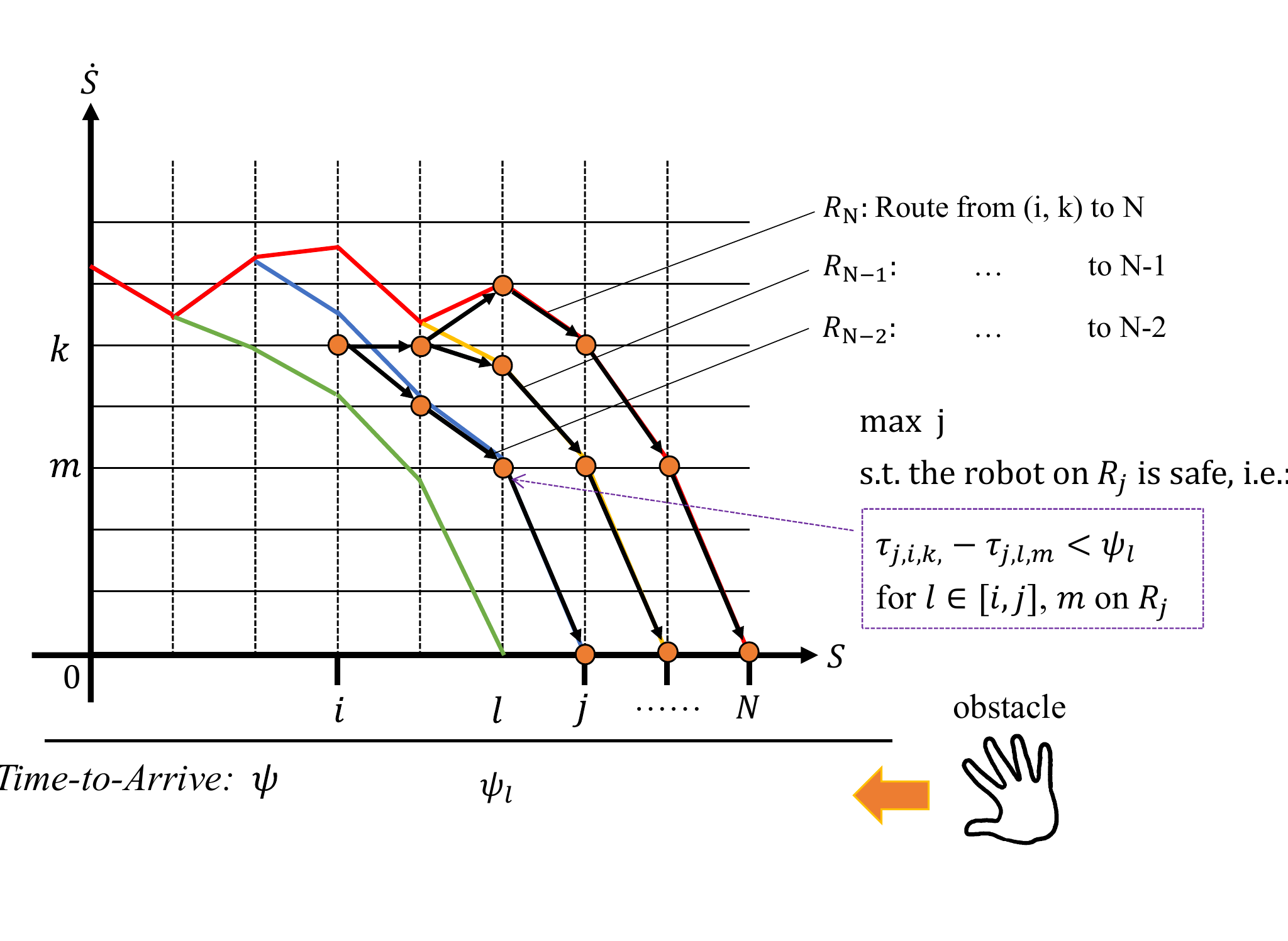}
  \vspace{-8mm}
  \subcaption{}
  \label{fig:method_execution}
  \end{center}
  \caption{Schematic illustrations of the pre-computation phase (a) and the execution phase (b).
    In the pre-computation phase, the algorithm computes \emph{stoppable sets} ((a)-left) backward from each stage, and compute \emph{Time-to-Reach} as a Dynamic Programming ((a)-right).
    In the execution phase, based on the current robot states and obstacle positions, the algorithm selects the best \emph{stoppable sets} and route, and calculates the fastest control inputs on every control cycle.
    }
  \label{fig:methods}
  \vspace{-2mm}
\end{figure}

This paper proposes a time-optimal control method where a robot moves the fastest, and yet safely, satisfying given constraints, founded on time-optimal path parameterization based on reachability analysis (TOPP-RA~\cite{phamNewApproachTimeOptimal2018}) and Dynamic Programming.
Before executing a trajectory, we scan the path and pre-compute \emph{Time-to-Reach}, the time to go from all the discretized states and stop at all the waypoints along the path.
Then, at runtime, using the computed \emph{Time-to-Reach}, we calculate the fastest velocity of the robot on every control cycle guaranteeing that the robot can stop anytime before collision happens (see \cref{fig:methods} for illustration).
As TOPP-RA handles generalized second-order constraints, our method inherits that characteristic and we can apply any second-order kinodynamic constraints, including joint velocities, accelerations, and torque constraints.
In addition, we also propose the method to accelerate the pre-computation of \emph{Time-to-Reach} dramatically by solving one-dimensional linear programming in parallel on GPU, which allows the whole pipeline to be executed at runtime.

This paper is organized as follows.
We survey related work on trajectory scaling and robot control under a dynamic environment in \cref{sec:related_work}.
Our time-optimal path tracking method is presented in \cref{sec:our_method}, followed by the explanation of the parallelized pre-computation phase on GPU.
In \cref{sec:experiment}, we evaluate our method in three sets of experiments.
First, we show that our control policy is less conservative than state-of-the-art safe robot control methods in a 1-D car simulation.
Then, the acceleration effect in the pre-computation phase by the parallelization on GPU is examined.
Finally, we demonstrate a real-time control of a 6 DoF robot in simulation where the robot moves back and forth between two positions and safely stops before colliding with randomly-moving obstacles.
In \cref{sec:conclusion}, some concluding remarks are made with some directions for future work.
In Appendix \ref{sec:proof}, we prove the time-optimality of our method.

\section{Related Work} \label{sec:related_work}
Trajectory planning decomposition into path planning and velocity planning under dynamic environments dates back to~\cite{kantEfficientTrajectoryPlanning1986}.
Later, the concept of trajectory scaling was proposed along with collision reaction strategy \cite{haddadinCollisionDetectionReaction2008}.

Various controllers have been proposed in the context of SSM and PFL.
As briefly mentioned in \cref{sec:intro}, methods for SSM  compute the maximal, safe velocity based on current position and (relative) velocity information~\cite{zanchettinSafetyHumanrobotCollaborative2016,ragagliaSafetyawareTrajectoryScaling2015}.
The problem is they do not fully exploit the entire path information and leverage the capability of robots enough.
Methods for PFL compute a maximal, safe velocity along a path by calculating `effective mass' or `apparent mass' of a robot manipulator, whose accompanying force is within the limits of ISO standards~\cite{slothComputationSafePath2018, herbsterNewApproachEstimate2021}.
Recently, a method to combine SSM and PFL has been proposed which allows some force/torque when collision happens to maximize the productivity and efficiency~\cite{lucciCombiningSpeedSeparation2020}.

To mitigate the conservative behavior of the controllers, human's reachable space prediction has been investigated ~\cite{ragagliaSafetyawareTrajectoryScaling2015,pereiraCalculatingHumanReachable2017,ragagliaTrajectoryGenerationAlgorithm2018}.
Obstacle motion prediction enlarges the configuration space that is expected to be collision-free and can improve the speed and productivity of collaborative robots.
Another attempts are to determine dynamic separation distance more precisely.
\cite{vicentiniTrajectorydependentSafeDistances2014} computes a trajectory-dependent separation distance to improve the reaction time.
\cite{glogowskiExtendedCalculationDynamic2019} calculates distances from the entire kinematic chain of the robot to a human, and controls the robot's speed adaptively.
However, those approaches are based on velocity information of humans/obstacles.
Considering that the maximum acceleration of hand motion can be up-to 6 \si{m/s^2}~\cite{app12083863}, the human's motion prediction is intrinsically difficult with external, cost-effective sensors.
Instead, we assume that only the positions of obstacles and their maximum velocities are given and the obstacles can change their direction quickly keeping their maximum speed.
Please note that human motion prediction can be integrated with our method.

Aside from collaboration and safety, time-optimal path parameterization(TOPP) has been investigated.
TOPP is a problem to find the fastest traversal time for a robot to track a given trajectory satisfying constraints.
One approach is to apply numerical integration to obtain the maximum and minimum acceleration profile~\cite{doi:10.1177/027836498500400301,pham_toppni}.
Another approach is to transform TOPP into a convex optimization problem~\cite{verscheure2008practical, doi:10.1177/0278364914527855}.
Recently, reachability-analysis based approach is invented in \cite{phamNewApproachTimeOptimal2018}, where it converts TOPP into a set of Linear Programming.
This method is well-known to be simple, efficient, and robust, and our method is built on top of this work.

\section{Time-Optimal Path Tracking with Safety Guarantees}\label{sec:our_method}

\subsection{Terminologies and Definitions}
Our method originates from TOPP-RA~\cite{phamNewApproachTimeOptimal2018}.
First, we briefly describe the terminologies and definitions in TOPP-RA that we use in this paper.
Consider an $N$-DoF robot system, whose configuration is represented as $q \in \mathbb{R}^n$.
A geometric path $P = q(s)_{s \in [0, s_{end}]}$ in the configuration space is given where $q(s)$ is a piece-wise $C^2$-continuous.
A \emph{time parameterization} is a piece-wise $C^2$ increasing scalar function $s(t): [0, T] \rightarrow [0, s_{end}]$.

As in~\cite{phamNewApproachTimeOptimal2018}, we discretize the interval $[0, s_{end}]$ into $N$ segments and $N+1$ stages:
\begin{equation}
  0 =: s_0, s_1, ..., s_{N-1}, s_N := s_{end}
\end{equation}
Let $u_i$ be a constant path acceleration over the interval $[s_i, s_{i+1}]$ and by $x_i$ the squared velocity $\dot s_i^2$ at the $i$ stage, the following relation is derived through algebraic operations:
\begin{equation}
	\begin{aligned}
		x_{i+1} = x_i + 2 \triangle_i u_i,\ \ \ \  i = 0...N-1
	\end{aligned}
\end{equation}
where $\triangle_i = s_{i+1} - s_i$.
The generalized constraints in a discretization scheme are considered in a time-parametrization technique:
\begin{equation} \label{eqn:generalized_constraints}
	\mathbf{a_i} u_i + \mathbf{b_i} x_i + \mathbf{c_i} \in {\mathscr C}_i
\end{equation}
where coefficients and terms are derived through manipulator's kinamatics/dynamics equations and constraints at each stage.

We also borrow the ideas of the $i$-stage set of \emph{admissible} control-state pairs
$\Omega_i := \{(u, x) | \mathbf{a_i} u + \mathbf{b_i} x + \mathbf{c_i} \in {\mathscr C}_i\}$,
i-stage \emph{admissible controls} $\mathcal U_i(x) := \{u | (u, x) \in \Omega_i \}$.
Let $\mathcal I$ a set of states, the ideas of i-stage one-step set $\mathcal Q_i(\mathcal I)$
and $i$-stage \emph{controllable set} $\mathcal K_i(\mathcal I_N) := \mathcal Q_i(\mathcal K_{i+1}(\mathcal I_N))$ are adopted as well.
Intuitively, \emph{admissible} means, given a state $x_i$, the control input $u_i$ satisfies the constraints \cref{eqn:generalized_constraints} and can be applied to the system, whereas
\emph{controllable} means there exists control(s) to steer the robot from its state towards the goal state(s) under constraints.
See \cite{phamNewApproachTimeOptimal2018} for details.

\subsection{Problem Formulation} \label{subsec:problem_formulation}
The problem we solve here is to: \emph{``compute the time-optimal control for the robot to track a path under constraints, while the robot must stop when it is not safe''}.
Specifically, following the concept of Speed and Separation Monitoring (SSM) in ISO standards~\cite{ISOTS15066}, ``safety'' in this paper is defined as ``the robot is static when the minimum distance between the robot and obstacles is under a given protective distance''.
In other words, a human can touch the robot only when it stops.

We make the following assumptions:
\begin{itemize}
	\item A geometric path $P$ to track is given.
	\item The minimum distances between a robot and all dynamic obstacles are monitored, or the field of view of sensors is large enough to detect unknown obstacles can be safely handled by the robot.
	\item Maximum velocities of dynamic obstacles are given.
\end{itemize}
Note that the modeling of obstacles is not a requirement.
Raw sensor data, such as point cloud, can be used as long as the real-time computation of the minimum distance between a robot model and captured points on dynamic obstacles is possible.
Additionally, \emph{if} you have an estimation of obstacles' motions, you can incorporate it in \cref{eqn:time_to_arrive}, which will produce a less-conservative motion.
However, in this paper, we do not trust and rely on such predictions as discussed in \cref{sec:related_work}.

Our technique can be separated into a pre-computation phase and an execution phase.
In the pre-computation phase, we compute \emph{controllable sets} for a robot to stop at each stage.
We call these sets as \emph{stoppable sets}.
In the execution phase, we select the best \emph{stoppable sets} among the set of \emph{stoppable sets} to go as farther as possible, and compute the time-optimal control input.
Note that the pre-computation phase takes only about $0.5$ \si{s} practically (see \cref{subsec:robot_demo}), such that the whole pipeline can be executed in near real-time.
We explain these two phases step-by-step in the following sections.

\subsubsection{Pre-computation Phase} \label{subsubsec:precomputation_phase}
\cref{fig:method_precomputation} illustrates the pre-computation phase.
Firstly, we compute stoppable sets $\mathcal K_{j}$ for each stage $j \in [1, N]$ (the left side of \cref{fig:method_precomputation}).
This step is almost equivalent to ``backward pass'' in TOPP-RA~\cite{phamNewApproachTimeOptimal2018}, but the difference is that we process the backward pass from each stage.
The stoppable sets are drawn in different colors in \cref{fig:method_precomputation}.

Secondly, given the number of discretization in velocity space $M$, we determine the discretization size of $\dot s$ with the following equation, used later for dynamic programming:
\begin{equation}
  \delta v := \frac{\sqrt{\max{\mathcal K}}}{M}
\end{equation}
where $\mathcal K$ is the union of all the stoppable sets $\mathcal K_{j}$ (which is actually $\mathcal K_N$, see \cref{eqn:allstoppable_sets_includedby_KN}).
Please note that we discretize the velocity space uniformly, not the $x$ i.e. the squared $\dot s$.

Then, for each stoppable set, we compute the fastest travel time $t_{travel,j,i,k}$, which is the fastest time to control the robot at stage $i$ with the velocity $k\delta v$ to the next $(i+1)$ stage under contraints, for $i \in [0, j-1], k \in [0, M]$ (see the right side of \cref{fig:method_precomputation}).
This step is equivalent to ``forward pass'' in TOPP-RA, but we compute the greedy controls for all discretized states $s$ and $\dot s$ for each stoppable set.
\begin{equation}
	\begin{aligned}
	u^{*} &:= {\max u} \textrm{,  s.t.: }  \\
		x_k + 2\triangle_iu &\in \mathcal K_{j, i} \textrm{, } u \in \Omega_i, x_k = k\delta v \label{eqn:1dlp}
	\end{aligned}
\end{equation}
\begin{align}
	x^{*}_{k, i+1} &:= x_k + 2\triangle_i u^{*} \\
	t_{travel,j,i,k} &= \frac{2\triangle_i}{\sqrt{x^{*}_{k, i+1}} + \sqrt{x_k}}
\end{align}
The complexity of this computation is $O(N^2 M)$ and the bottleneck is 1-D Linear Programming of \cref{eqn:1dlp}.
We propose a parallelization technique for accelerating this step in \cref{subsec:parallel_1dlp}.

Finally, we apply dynamic programming to compute \emph{Time-to-Reach} $\tau_{j, i, k}$, which is the fastest time to go from the starting stage $i$ at the velocity $k\delta v$ and stop at the stage $j$.
We also record the fastest travel $ \rho_{j,i,k} $ for each grid point.
This processing enables parallelization at runtime execution phase.
\begin{equation}
	\begin{aligned}
    \rho_{j,i,k} &= \bigl\lfloor(\frac{\sqrt{x^{*}_{k, i+1}}}{\delta v})\bigr\rfloor \\
		\tau_{j, j, all} &= inf, \ \ \tau_{j, j, 0} = 0 \\
		\tau_{j, i, k} &= \tau_{j, i+1, \rho_{j,i,k}} + t_{travel,j,i,k}
	\end{aligned}
\end{equation}

The overall algorithm is described in \cref{alg:time_to_reach_dp}.
Note that floor $\lfloor\rfloor$ and ceil $\lceil\rceil$ operations are used in calculating velocity indices in a ``safe'' manner, i.e.,  not violating constraints and not over-estimating the fastest travel time.

Now we have \emph{Time-to-Reach} $\tau$ and the fastest travel $\rho$, used in the execution phase for real-time control explained in the next section.

\subsubsection{Execution Phase} \label{subsubsec:execution_phase}
\cref{fig:method_execution} illustrates how to compute the time-optimal controls.
Let $d_{protective}$ denote a given protective distance.
Consider that the robot is at the position of the stage $i$ with the velocity $v_i$.

First, we compute the minimum distances $d_{l,q}$ between a robot that locates at every stage $l$ and each obstacle $q$, and then calculate the \emph{Time-to-Arrive} $\psi_l$ for each $l$ stage:
\begin{equation} \label{eqn:time_to_arrive}
	\psi_l = \min_{q}\frac{d_{l, q} - d_{protective}}{v_{q, max}}
\end{equation}
where $v_{q, max}$ denotes the maximum velocity of obstacle $q$ in Cartesian space.

To control a robot to go as fast as possible, we select the farthest stage to stop, where the robot stops and then (possibly) collides:
\begin{equation} \label{eqn:collision_selection}
  \begin{aligned}
    j_{stop} &= \max_{j \in [i, N]}{j} \\
    \textrm{ s.t.: } (\tau_{j,i,k} - \tau_{j,l,m}) &< \psi_j \textrm{   } \forall l \in [i, j], \textrm{m on the route } R_j
  \end{aligned}
\end{equation}
where the stopping route $R_j$ can be obtained by tracing the fastest travel $\rho_j$.

Finally, we apply the exact ``forward pass'' of TOPP-RA from $i$ to $j_{stop}$ to compute the time-optimal path parameterization using the stoppable set $\mathcal K_{j_{stop}}$, and apply the fastest velocity $s_i$ at the stage $i$ to the robot.
We repeat this execution process for every control cycle.
Practically, this runs within $6$ milliseconds for $500$ stages in a 6 DoF robot experiment in \cref{subsec:robot_demo}, which is suitable for real-time control.

The farthest stage calculation has a computational cost of $O(N^2)$, but it is computationally fast enough, typically completing in 1--2 \si{\milli \second} at most.
In cases where longer trajectories exist and/or fine discretization is required, we can parallelize this process owing to the formulation with \emph{Time-to-Reach}.
The parallelized version enumerates all the routes to be traced beforehand, and evaluates \cref{eqn:collision_selection} for every iteration in a batched and scalable manner.
We observed that the parallelized algorithm runs in less than 1 \si{\milli \second} at worst on GPU, with additional costs primarily for memory allocation and data transfer during the pre-computation phase.

In Appendix \ref{sec:proof}, we show that our method is time-optimal in the sense that
\emph{For any robot motion that is strictly faster than the motion recommended by our policy, there exists a human motion that results in a collision with the robot in a non-stationary state.}
See the Appendix \ref{sec:proof} for the proof.

\begin{algorithm}[tbhp]
    \caption{Compute Time-to-Reach as a Dynamic Programming}
    \label{alg:time_to_reach_dp}
    \begin{algorithmic}[1]  %
      \STATE Given $M$: the number of discretization in $x$
      \STATE Output {$\mathcal K$ : Stoppable sets - a set of Controllable sets $\mathcal K_{j}$ for stopping at stage $j$, where $j \in [0, N+1]$},
		\STATE Output {$\tau_{j, i, k}$ : Time-to-Reach starting from the stage $i$ and the velocity index $k$, and stopping at stage $j$, where $j \in [1, N], i \in [0, N-1], k \in [0, M]$}
      \STATE
      \STATE \textup{/* Backward pass for stopping sets */}
      \FOR{$j \in [N, 1]$}
        \STATE $\mathcal K_{j, N} := {0}$
        \FOR{$i \in [N-1, 0]$}
          \STATE $\mathcal K_{j, i} := \mathcal Q_i(\mathcal K_{j, i+1})$
        \ENDFOR
      \ENDFOR
      \STATE $\delta v := \frac{\sqrt{\max{\mathcal K}}}{M}$
      \STATE \textup{/* For each stoppable set */} \label{alg:line:afterstopsets}
      \FOR{$j \in [N, 1]$}
        \STATE \textup{/* Dynamic Programming */}
        \STATE $\tau_{j, j, {all}} := \inf$
        \STATE $\tau_{j, j, 0} := 0$
        \FOR{$i$ in $[j-1, 0]$}
          \STATE $x^{-}_i, x^{+}_i \leftarrow \mathcal K_{j, i}$
		  \STATE $l := \Bigl\lceil\frac{\sqrt{x^{-}_i}}{\delta v}\Bigr\rceil$, $m := \Bigl\lfloor\frac{\sqrt{x^{+}_i}}{\delta v}\Bigr\rfloor$
          \FOR{$k$ in $[l, m]$}
            \STATE /* Compute the fastest travel time at stage $i$ */
            \STATE $u^{*} := {\max u}$,  s.t.: $x_k + 2\triangle_iu \in \mathcal K_{j, i}$, $x_k = (k\delta v)^2$, $u \in \Omega_i$
            \STATE $x^{*}_{k, i+1} := x_k + 2\triangle_i u^{*}$
            \STATE $v := \sqrt{{x_k}}$,  $ v^{*}_{k, i+1} := \sqrt{x^{*}_{k, i+1}}$
            \STATE $t = 2\triangle_i / (v^{*}_{k, i+1} + v)$
			\STATE \textup{/* Compute Fastest Travel and Time-to-Reach */}
            \STATE $\rho_{j,i,k} = \bigl\lfloor(\frac{v^{*}_{k, i+1}}{\delta v})\bigr\rfloor$
            \STATE $\tau_{j, i, k} := \tau_{j, i+1, \rho_{j,i,k}} + t$
          \ENDFOR
        \ENDFOR
      \ENDFOR
\end{algorithmic}
\end{algorithm}

\subsection{Parallel 1-D Linear Programming on GPU} \label{subsec:parallel_1dlp}
Our primary contribution is that we accelerate the pre-computation phase computation by 10x, which allows the whole pipeline to be executed at runtime.
The key idea is the parallelization of 1 Dimensional Linear Programming (LP) (\cref{eqn:1dlp}) on GPU.
The 1-D LP can be transformed into the following form:
\begin{equation}
	\begin{aligned}
		\max\ & {u} \\
		s.t.\ & a_i u \leq b_i,  a_i \neq 0
	\end{aligned}
\end{equation}
Note that the notations $a_i, b_i$, and $c$ are re-used in different meanings from \cref{eqn:generalized_constraints}.
This can be easily solved as follows~\cite{millerLecture16Linear}.
Constraints are classified into two groups - two half-spaces $C^{+}$, $C^{-}$:
\begin{equation}
	\begin{aligned}
		C^{+} &= \{i | u \leq b_i / a_i\} \\
		C^{-} &= \{i | u \geq - b_i / a_i\} \\
	\end{aligned}
\end{equation}
Then, when $\alpha \leq \beta$ where $\alpha := \max{\{-b_i/a_i|i \in C^{-}\}}$ and $\beta := \min{\{b_i/a_i | i \in C^{+}\}}$, the 1-D LP is feasible and its optimal solution $u$ can be found as:
\begin{equation}
	\begin{aligned}
		u^{*} = \beta
	\end{aligned}
\end{equation}

Here, we only use basic arithmetic operations and $\max, \min$ whose performance for parallel computation is pretty optimized in GPGPU, and computationally-efficient APIs are provided by GPGPU frameworks such as PyTorch~\cite{paszke2017automatic}.
No if/else switching is required that causes warp divergence leading to poor performance~\cite{hanReducingBranchDivergence2011}.

\section{Experiments and Results} \label{sec:experiment}
\subsection{Comparison with Existing Method} \label{subsec:algorithm_performance}
\begin{figure}[thbp]
  \begin{center}
  \includegraphics[width=0.9\hsize,trim={0cm 0.8cm 0.3cm 1.1cm},clip]{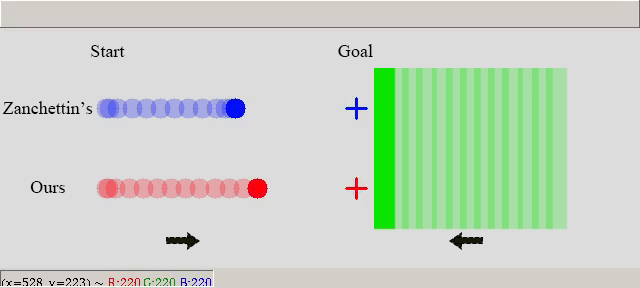}
  \caption{1-D car simulation setting for comparison}
  \label{fig:1d_experiment_setting}
  \end{center}
\end{figure}

Firstly, we compare our method with the existing state-of-the-art method (Zanchettin's) \cite{zanchettinSafetyHumanrobotCollaborative2016} in a simple 1-D car simulation~(\cref{fig:1d_experiment_setting}).
We put two cars at the same position (0 \si{m}) and they start to move from left to right at the same time towards the same goal position (25 \si{m}) controlled by each method.
The car above represents Zanchettin's method and the below one is ours.
Cars have the same velocity and acceleration limits (20 \si{m/s} and 100 \si{m/s^2} respectively).
For our method, the protective distance is set to 0.
A green wall moves from right to left to hinder the cars at its constant speed (20 \si{m/s}), and when it collides with the faster car, waits for 1 \si{sec}, and moves back to the right direction.
The stopping time in Zanchettin's method is manually optimized to (0.55 \si{s}) to be small enough not to have an infeasible optimization problem.

\begin{figure}[thbp]
  \begin{center}
  \includegraphics[width=1.0\hsize,trim={0.0cm 0.0cm 0.0cm 0.3cm},clip]{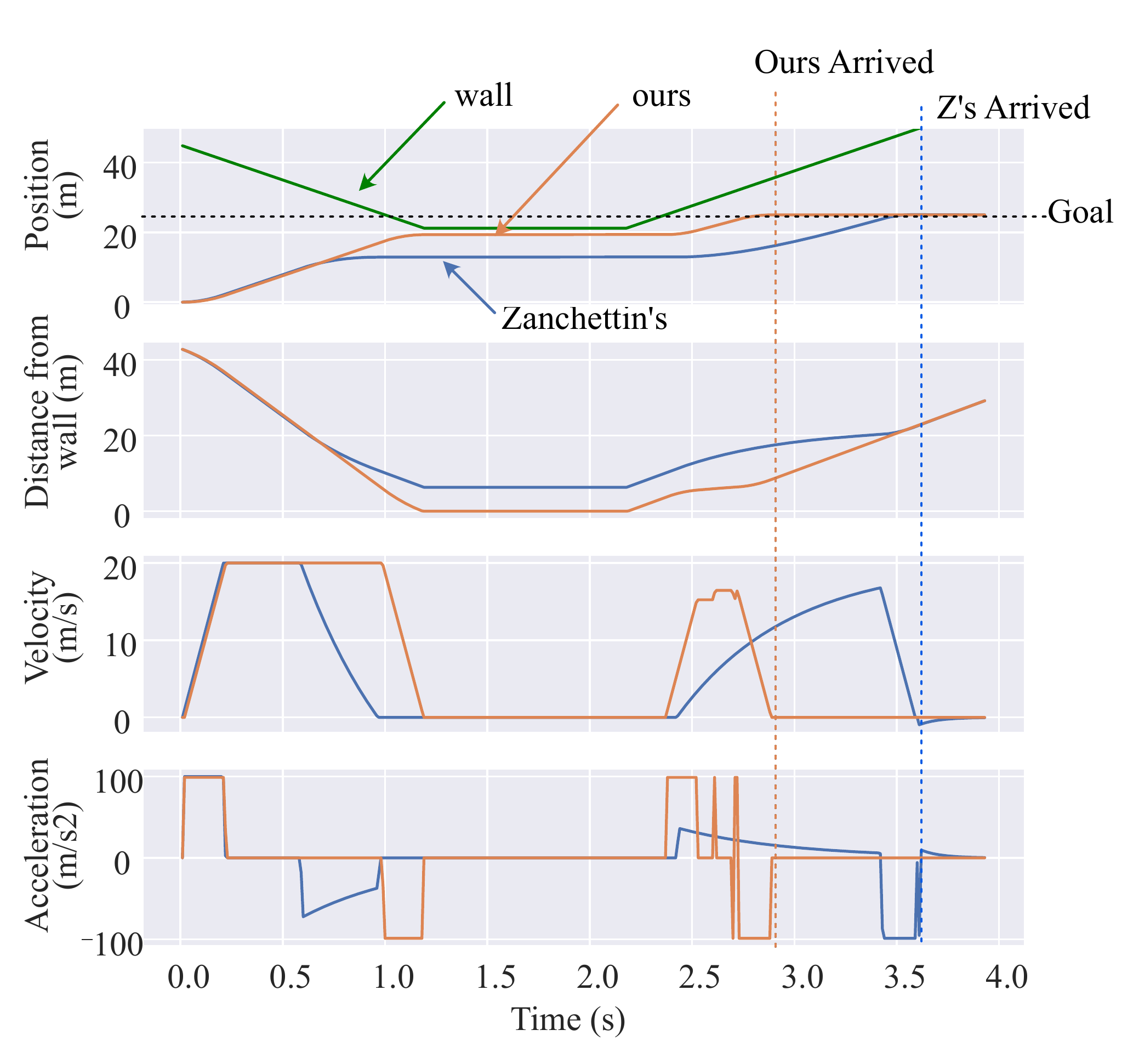}
  \caption{Comparison of Zanchettin's and Ours - Positions, Distance between cars and a wall, Velocities and Accelerations in a time series. See the video of this simulation at \url{https://youtu.be/SHwyOOU3X2A}.}
  \label{fig:1d_experiment_comparison_result}
  \end{center}
\end{figure}

We show the result in \cref{fig:1d_experiment_comparison_result}.
At first, two cars accelerate at their maximum speed.
Then, while the Zanchettin's car gradually decelerates and stops as the wall approaches way before the wall collides, ours stops almost exactly when the distance becomes 0 at its maximum deceleration.
As a result, our car arrives earlier than Zanchettin's.
The experiment can be viewed at \url{https://youtu.be/SHwyOOU3X2A}.

We summarize the following observations about Zanchettin's method from this experiment:
`Stopping time' needs to be large and conservative enough to consider the case when a robot and an obstacle come at their max speed.
Otherwise, the optimization problem is not always feasible e.g., robot cannot decelerate enough to stop in time.
This property conservatively constrains the velocity from the upper side with the constraint (9b) of the LP problem (9) in~\cite{zanchettinSafetyHumanrobotCollaborative2016}.
Moreover, the authors claim that `$\dot q_{k+1} = 0, \delta_k = 0$ is always a solution', but it is not true because, due to lower acceleration limits, the optimization problem can be infeasible when the robot moves too fast to decelerate enough within a stopping time to avoid unsafe collision.

\subsection{Parallel 1-D Linear Programming on GPU} \label{subsec:parallel_1dlp_performance}
\begin{figure}[thbp]
  \begin{center}
  \begin{minipage}[b]{0.8\hsize}
  \includegraphics[width=0.9\hsize,trim={0cm 0cm 1cm 0cm},clip]{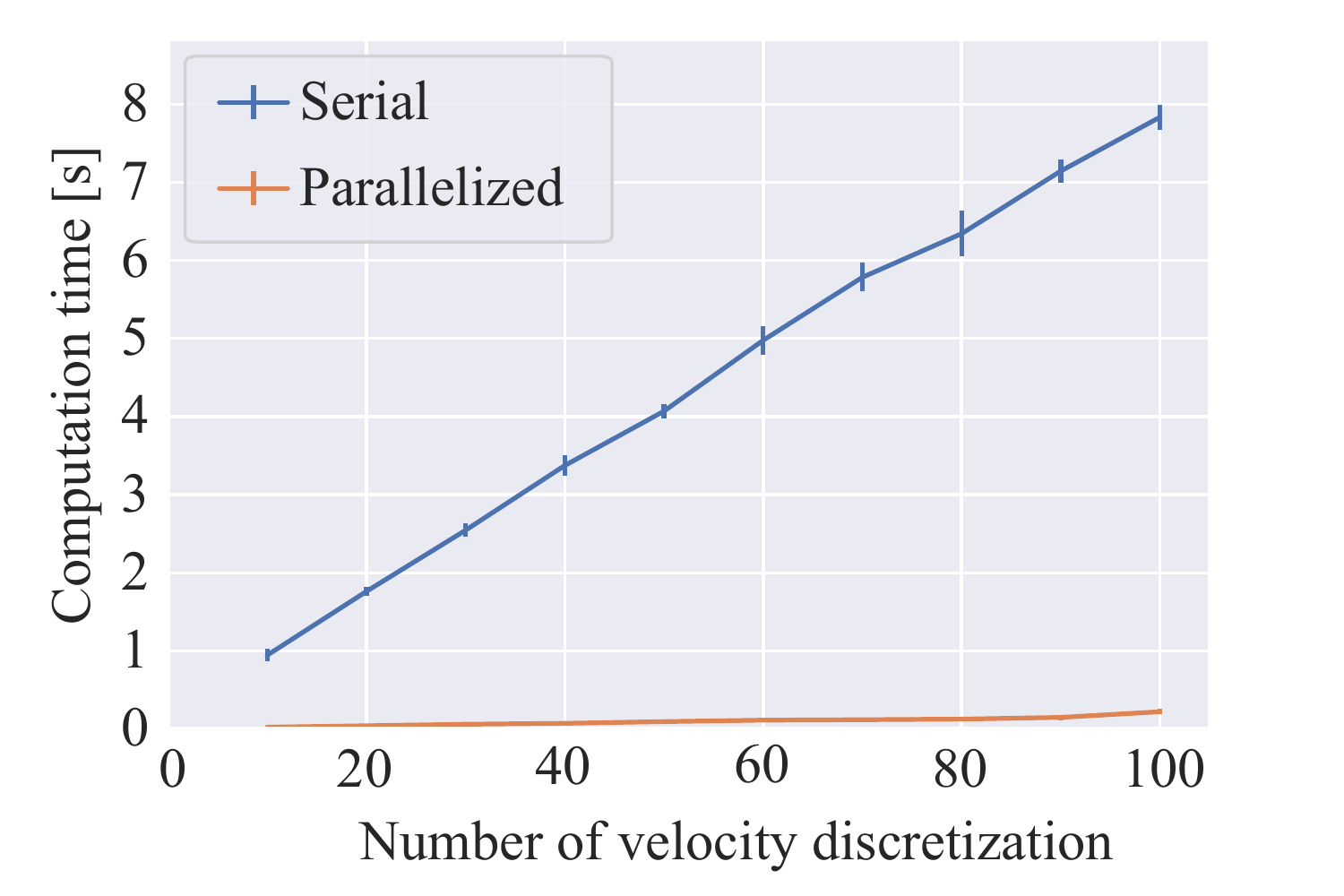}
  \subcaption{Effect of the number of velocity discretization $M$}
  \end{minipage}
  \begin{minipage}[b]{0.8\hsize}
  \includegraphics[width=0.9\hsize,trim={0cm 0.5cm 1cm 0cm},clip]{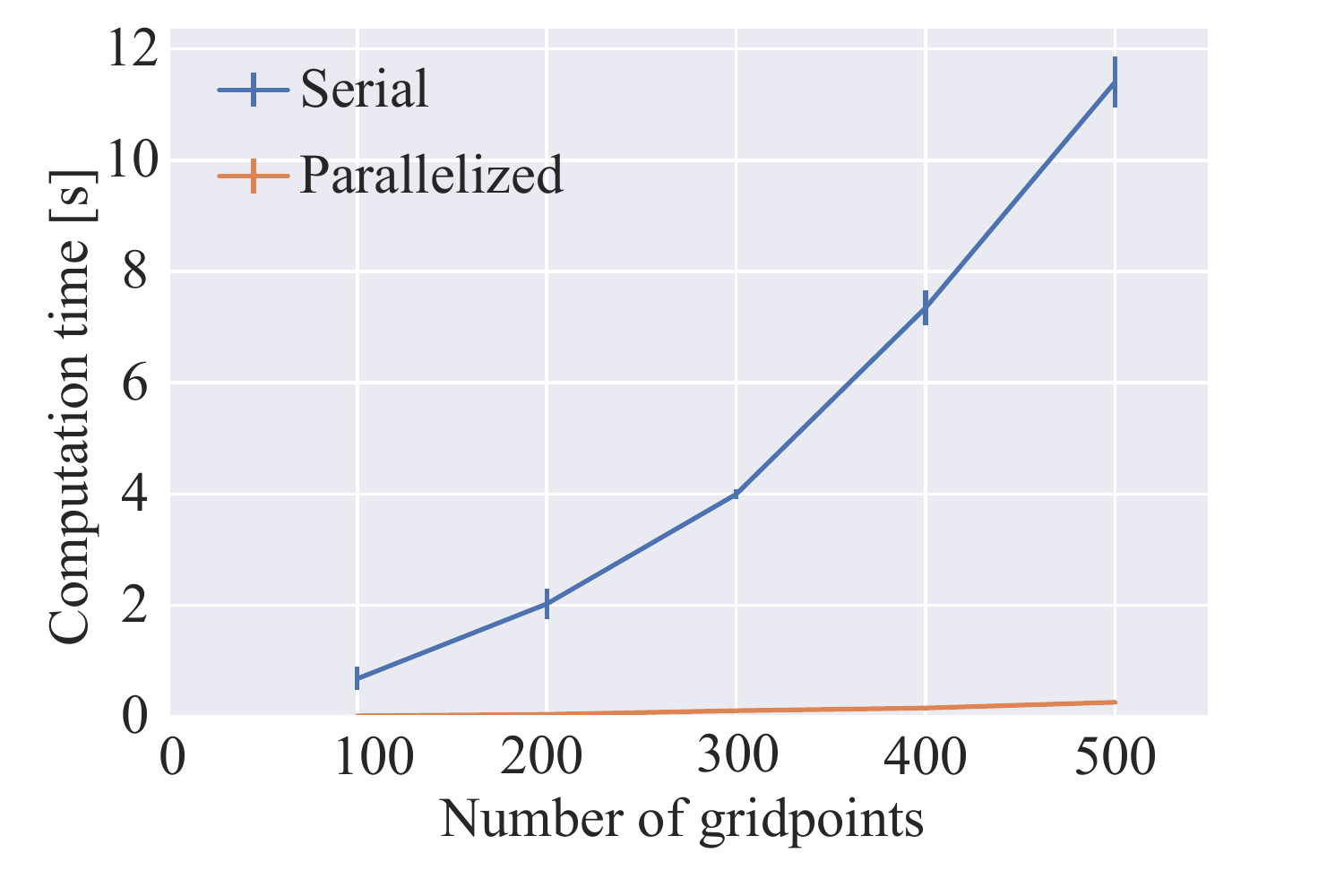}
  \subcaption{Effect of the number of stages $N$}
  \end{minipage}
  \caption{\emph{Time-to-Reach} computation time comparison, Serial processing v.s. our Parallelized 1-D Linear Programming solver.}
  \label{fig:parallel_1dlp_experiment}
  \end{center}
\end{figure}

Secondly, we experimentally show the performance of parallel 1-D Linear Programming solver running on GPU, compared to serial computation with vanilla TOPP-RA on CPU.
The performance is measured with the computation time from the line \ref{alg:line:afterstopsets} to the end of \cref{alg:time_to_reach_dp} that is right after the computation of stoppable sets $\mathcal K$ and includes Dynamic Programming, changing the number of stages $N$ and the number of velocity discretization $M$.
We randomly generate three 6-DoF waypoints, interpolate them with a spline, and compute its \emph{Time-to-Reach} under randomly generated joint velocity limits and acceleration limits for 10 trials.
Our method is implemented in Python with PyTorch, and the serial processing version is in Python with Cython.
The serial version is optimized to use memorization of the solutions of 1-D LP problems for the case when $\mathcal K_{j_1,i} \textrm{ and } \mathcal K_{j_2,i} \textrm{ } (j_1 \ne j_2)$ are the same. %
This experiment is executed on a single machine, whose CPU is Intel\textsuperscript{\textregistered} Xeon\textsuperscript{\textregistered} W-2145 and GPU is GeForce GTX 1080 Ti.
We fix $N = 300$ when we change $M$, and $M = 50$ when we change $N$.
\cref{fig:parallel_1dlp_experiment} clearly shows that the computation time increases linearly to $N$ and to the square of $M$.
The time for pre-computation with serial processing takes 1 \si{sec} even with the smallest numbers of discretization ($N=300, M=10$ and $N=100, M=30$) in this experiment, which is too coarse and useless for real applications.
On the other hand, our parallelized 1-D LP solver dramatically outperforms serial processing and takes only 0.25 \si{sec} even with the largest number of discretization in the experiment ($N=500, M=30$).
This result clarifies our contribution for allowing the whole pipeline to be adopted at runtime, nearly in real-time.

\subsection{Simulation on a 6-Dof Industrial Robot} \label{subsec:robot_demo}

\begin{figure}[thbp]
  \begin{center}
  \begin{minipage}[b]{1.0\hsize}
  \begin{center}
  \includegraphics[width=0.8\linewidth,trim={1cm 3cm 1cm 3cm},clip]{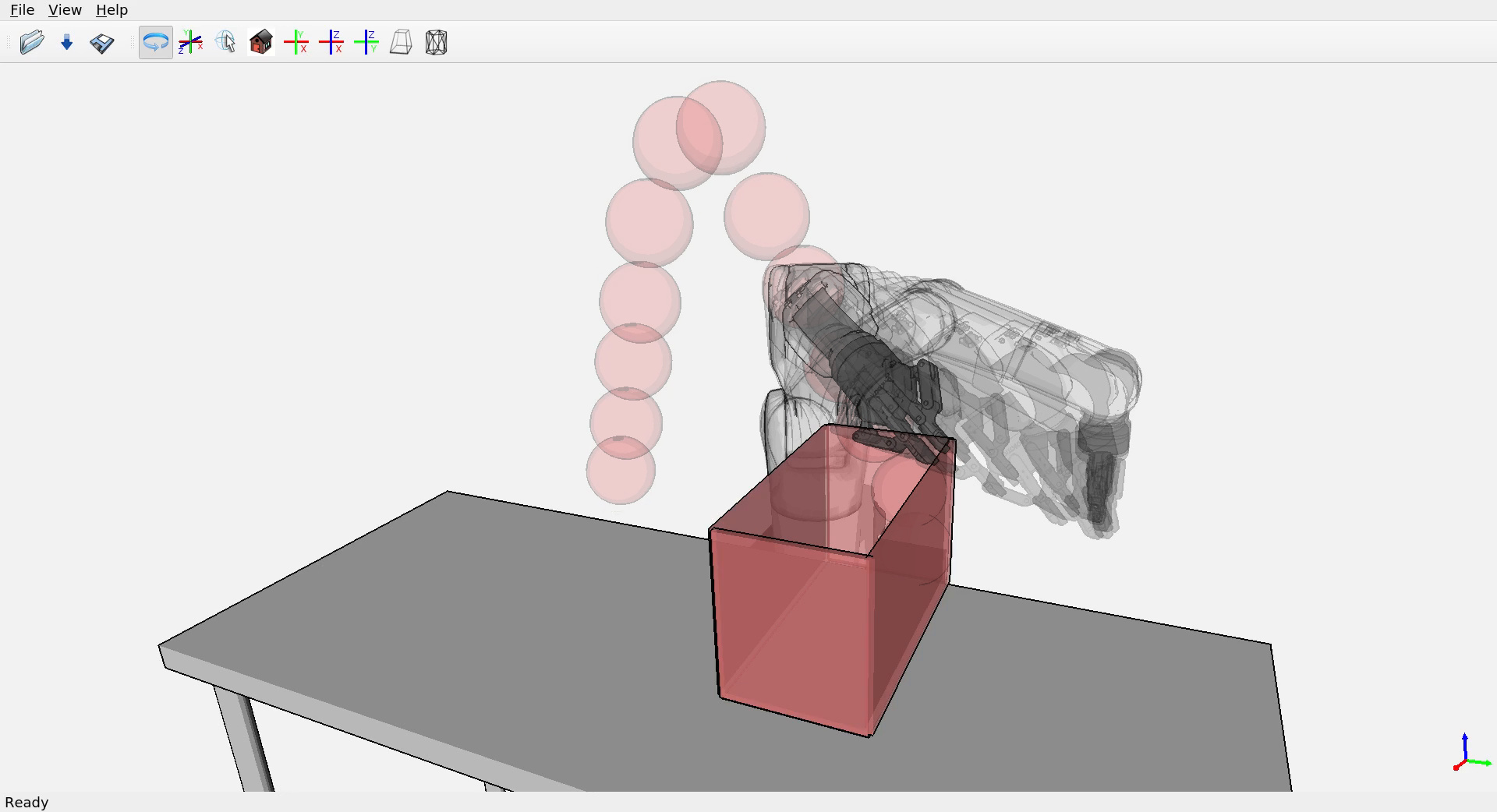}
  \subcaption{}
  \label{fig:robot_experiment_1obs}
  \end{center}
  \end{minipage}
  \vspace{1mm}
  \begin{minipage}[b]{1.0\hsize}
  \begin{center}
  \includegraphics[width=0.8\linewidth,trim={9cm 3cm 9cm 4cm},clip]{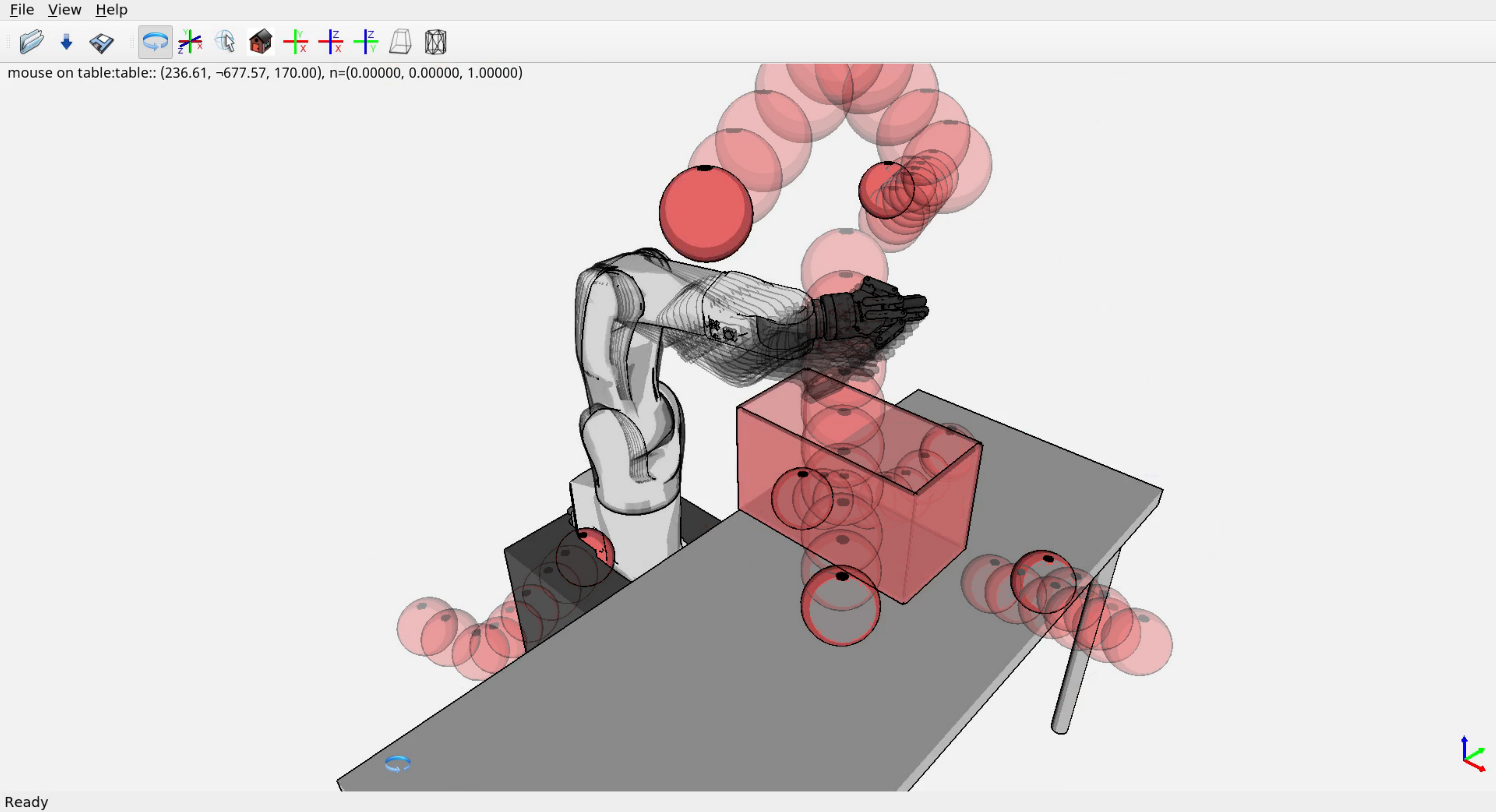}
  \subcaption{}
  \label{fig:robot_experiment_6obs}
  \end{center}
  \end{minipage}
  \end{center}
  \caption{(a) 6-DoF Robot control experiment in a simulation where the robot moves ``safely'' in an environment in which 1 dynamic obstacle is randomly moving at its maximum speed.
    (b) Our method can navigate the robot through a crowded environment with 6 randomly-moving dynamic obstacles.
    See the video of this experiment at \url{https://youtu.be/ta3lx80jJjk}}
  \label{fig:robot_experiment}
\end{figure}

\begin{figure*}[thbp]
  \begin{center}
  \includegraphics[width=1.0\linewidth,trim={1cm 0.0cm 3cm 0cm},clip]{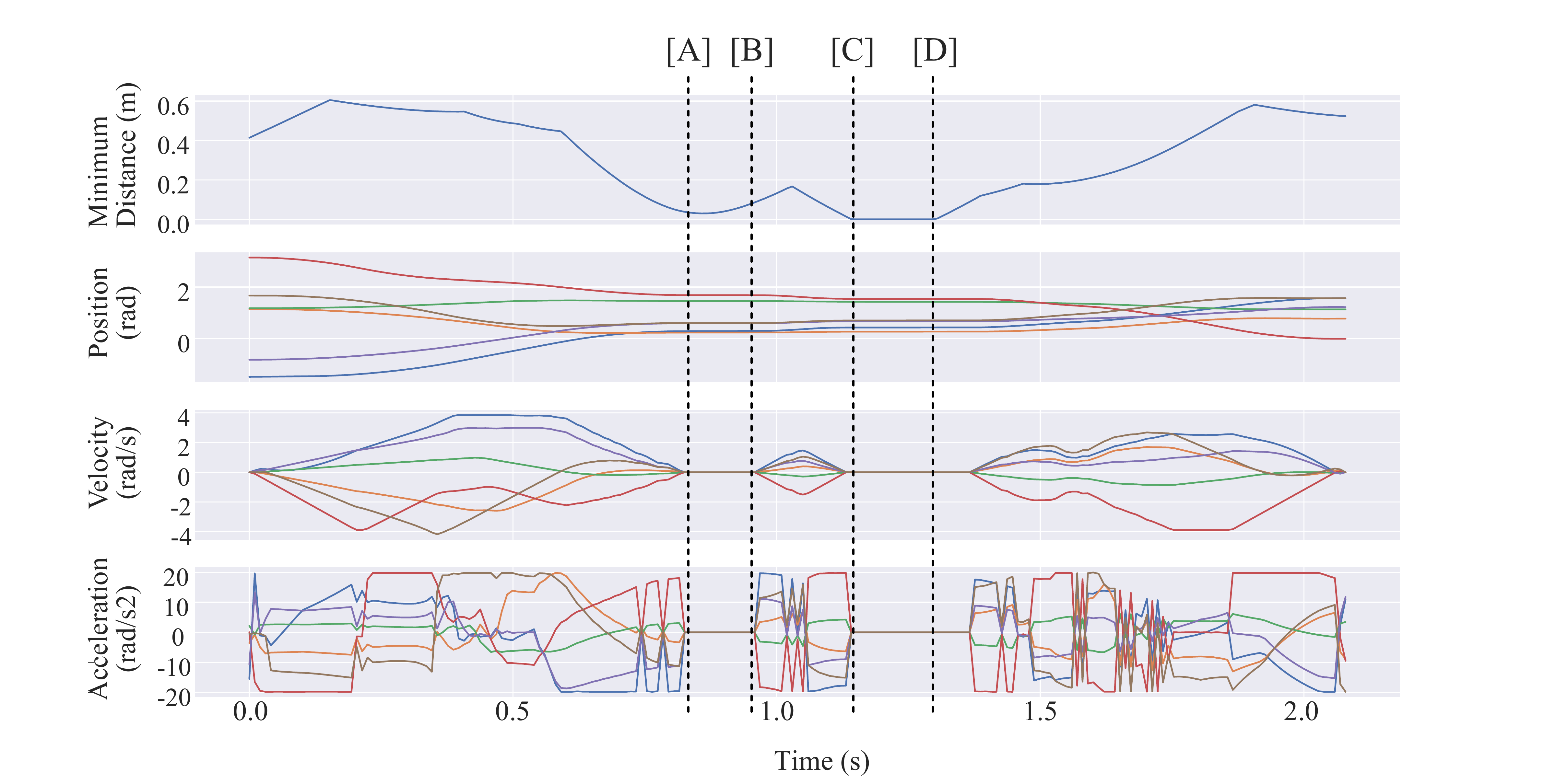}
  \caption{A result of one trajectory execution, logging Minimum distance from a dynamic obstacle, Joint Positions, Velocities and Accelerations.
      The robot decelerates at its maximum accelerations ($\pm20$ \si{rad/s^2}) and stops (velocity is 0) when the obstacle approaches ([A]--[B]) and collides ([C]--[D]) with the robot, and restarts its motion without violating joint velocity limits and acceleration limits.
	  }
  \label{fig:control_result}
  \end{center}
\end{figure*}

Finally, we conducted a 6 DoF robot experiment in simulation~(\cref{fig:robot_experiment}).
In \cref{fig:robot_experiment_1obs}, A robot moves back-and-forth between two positions while a dynamic obstacle moves in a bouncy, jig-zag manner in front of the robot at the speed of 1.6 m/s, the conservative human's speed as reported in~\cite{ISOTS15066}.
The protective distance is set to 0 for demonstration.
For fast minimum-distance computation, we simplify the robot by modeling all the robot links (not only the end-effector) with their enclosing spheres and compute the distances between the centers of the spheres and the center of the obstacle.
Our algorithm is implemented on top of OpenRAVE\cite{diankov_thesis} and the path the robot tracks is computed by an off-the-shelf Bi-RRT planner and a parabolic smoother provided by OpenRAVE.
The program runs on a laptop with AMD Ryzen 9 4900HS and NVIDIA GeForce RTX 2060 with Max-Q Design.
The number of velocity discretization $M$ is set to 30 and the minimum number of stages $N$ is set to 500, resulting in 517 stages proposed by TOPP-RA library~\cite{toppra_github}.
Throughout the experiment, the robot moves, stops safely at its maximum accelerations ($\pm20$ \si{rad/s^2}) when the obstacle approaches ([A]--[B] in ~\cref{fig:control_result}) and collides ([C]--[D] in ~\cref{fig:control_result}) with the robot, and restarts its motion without violating joint velocity limits and acceleration limits.
Our method works even in a cluttered environment where 6 dynamic obstacles move at 1.6 \si{m/s} in random directions~(\cref{fig:robot_experiment_6obs}).
The mean value of total pre-computation for 20 trials is accelerated from $4.03\pm1.01$ \si{s} to $0.40\pm0.09$ \si{s} by our proposed parallelized 1-D LP solver, which is 10 times faster.
Both results include the computation time of stoppable sets ($0.20\pm0.04$ sec) that runs on 16 CPU processes in parallel.
For reference, when enabling GPU at execution phase for farthest stopping stage computation, the pre-computation phase takes $0.54\pm0.16$ \si{s} due to additional cost for memory allocation and data transfer.
The total computation of the execution phase runs within $6$ \si{\milli \second} on CPU, and $5$ \si{\milli \second} on GPU.
The experiment can be viewed at \url{https://youtu.be/ta3lx80jJjk}. %

\section{Conclusion} \label{sec:conclusion}
In human-robot collaboration, the speed and the safety are generally in a trade-off relationship, which has led collaborative robots to be conservative and less productive.
This paper proposes a time-optimal control method for a robot to track a path guaranteeing the safety of human workers satisfying given constraints, based on TOPP-RA and Dynamic Programming, according to the SSM framework in ISO standards.
Our controller provides time-optimal control inputs at runtime and is strictly less conservative than the state-of-the-art controller guaranteeing collision safety, that is experimentally shown in simulation and is demonstrated in a 6-DoF robot experiment.
In addition, the pre-computation phase is accelerated by leveraging GPU and has been reduced down to 0.5 \si{s} in the 6 DoF robot experiment, which is 10x faster than that of vanilla implementation in TOPP-RA and can be executed at runtime nearly in real-time.

Future work can be in the following directions:
The computation of stoppable sets is a sequential process whose computational complexity is $O(N^2)$.
It does not fit parallel operations on GPU and now becomes a bottleneck of the entire pre-computation phase.
Another problem is space complexity.
Current implementation consumes about 5 GB memory on GPU in a robot experiment (\cref{subsec:robot_demo}), and in case we apply more constraints it will increase in $O(a^3)$ where $a$ is the number of coefficients of constraints.
Yet another remark is about the smoothness.
For example, we observe that, when the human is on the same trajectory of the robot and followed by the robot, the robot repeatedly accelerates and decelerates to come to the next waypoint as fast as possible once the human passes a waypoint, resulting in a ``bang-bang'' acceleration profile (we can see the spikes in \cref{fig:1d_experiment_comparison_result}).
However, it would be desirable for the robot to smoothly track its trajectory and follows the human, even if it sacrifices ``time optimality''.
Exploring this \emph{human-aware} path tracking is an interesting future direction.

\begin{appendices}
\section{Proof of Optimality} \label{sec:proof}
First, we prove the following theorem:

\emph{Theorem 1: }
The stoppable sets $\mathcal K_{j,i}$ always satisifies the following:
\begin{equation}
	\mathcal K_{j,i} \subseteq \mathcal K_{k,i} \textrm{ s.t.  } \forall j < k, \forall i \leq k
\end{equation}

\begin{figure}[thbp]
  \begin{center}
  \includegraphics[width=1.0\linewidth,trim={8cm 4cm 6cm 6.2cm},clip]{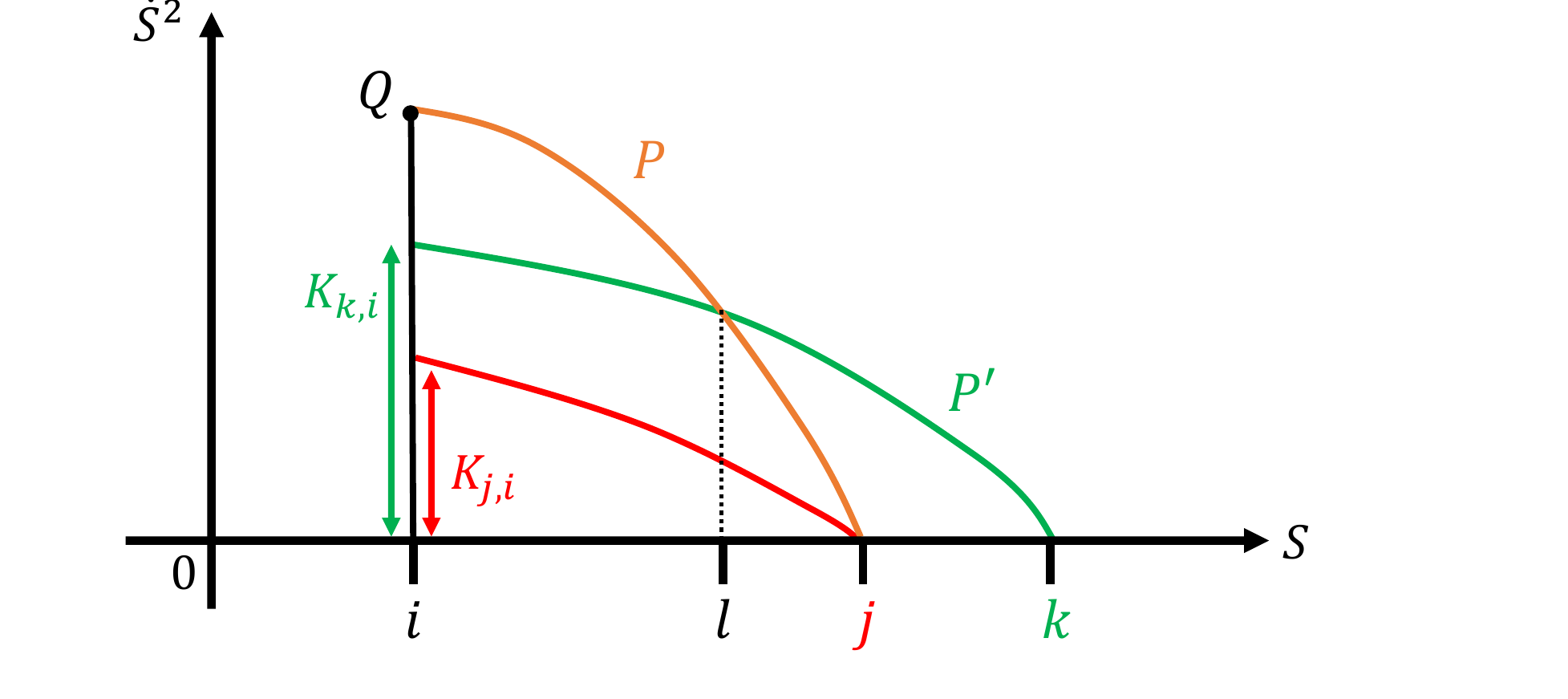}
  \end{center}
  \caption{A figure for the proof of inclusion relationship between the stoppable sets.}
  \label{fig:proof1}
\end{figure}

\begin{figure}[thbp]
  \begin{center}
  \begin{minipage}[b]{0.8\hsize}
  \begin{center}
  \includegraphics[width=1.0\linewidth,trim={9cm 4cm 9cm 4.8cm},clip]{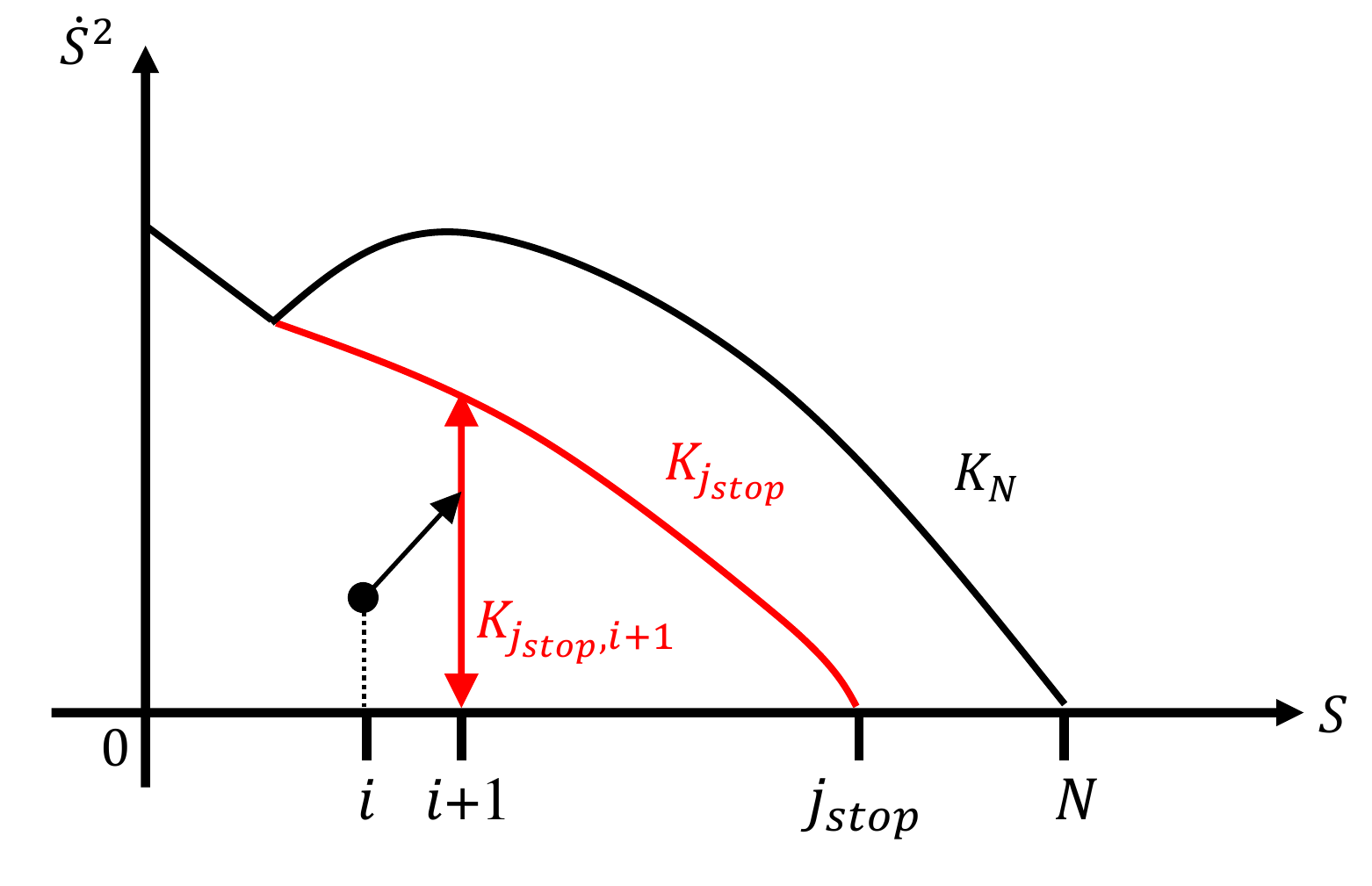}
  \subcaption{}
  \label{fig:proof2a}
  \end{center}
  \end{minipage}
  \begin{minipage}[b]{0.8\hsize}
  \begin{center}
  \includegraphics[width=1.0\linewidth,trim={9cm 3cm 9cm 4.8cm},clip]{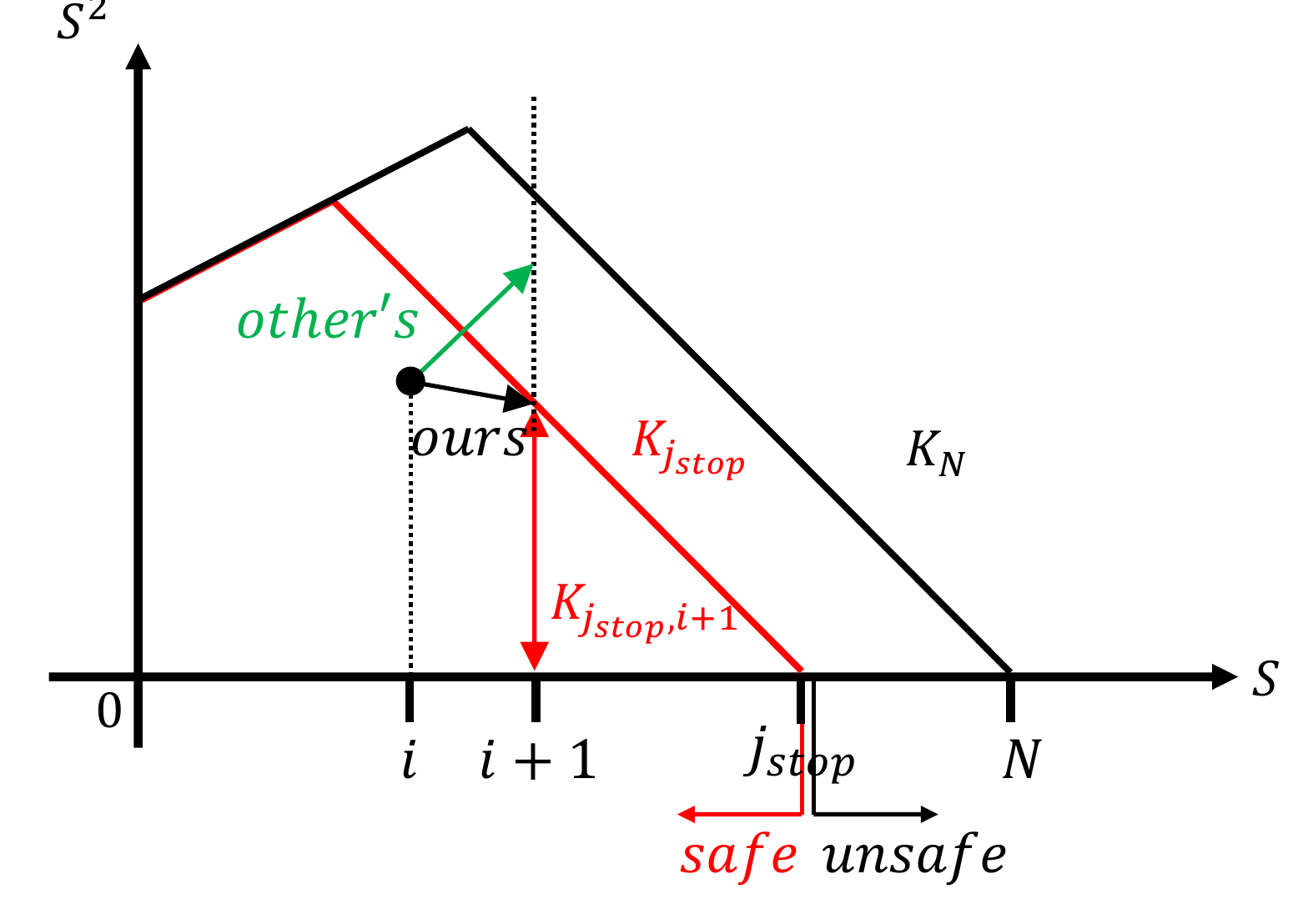}  %
  \subcaption{}
  \label{fig:proof2b}
  \end{center}
  \end{minipage}
  \vspace{-3mm}
  \end{center}
  \caption{Figures for the proof of the optimality of our method.}
  \label{fig:proof_figures}
\end{figure}

\emph{Proof:}
Assume by contradiction that there exists a velocity $\dot s^2$ that is $\mathcal K_{j, i}$ but not in $\mathcal K_{k, i}$ (the point $Q$ in \cref{fig:proof1}).
By definition of $\mathcal K_{j, i}$, there exists a profile $P$ (the orange line) starting from that $\dot s^2$ ($Q$) and that stops at $j$.
By continuity of the boundaries (see the Appendix A of TOPP-RA paper~\cite{phamNewApproachTimeOptimal2018}), there exists a time $l$ where the profile intersects the boundary of $\mathcal K_k$ (the green profile).
We construct a profile $P'$ by glueing toghther the beginning of the orange profile (P) and the end of the green profile.
We have then proved that $\dot s^2 \in \mathcal K_{k, i}$, which contradicts the initial assumption. $\square$

From Theoreom 1, the following can be derived:
\begin{equation} \label{eqn:allstoppable_sets_includedby_KN}
    \mathcal K_{j,i} \subseteq \mathcal K_{k,i} \ldots \subseteq \mathcal K_{N,i} \textrm{,  } \forall i
\end{equation}

Based on the above, we derive the following theorem:

\emph{Theorem 2: }
For any robot motion that is strictly faster than the motion recommended by our policy,
there exists a human motion that results in a collision with the robot in a non-stationary state.

\emph{Proof:}
Suppose that the robot is at $i$, our method selects the stoppable set $\mathcal K_{j_{stop}}$ and computes a control at stage $i$ using TOPP-RA's one-step greedy forward pass.
Assume by contradiction that there exists a strictly faster and safe control $u^{**}_i$ by other motion policy.

Since the robot must be controlled to stop at the final destination ($s_{N}$), any motion policies keep the robot state $s^2$ satisfies $\dot s^2_l \in \mathcal K_{N} \textrm{ s.t. } (i < l \leq j_{stop} \leq N) $.
In addition, any control input driven by all the other policies must also be \emph{admissible}, i.e., $u \in \mathcal U_i(x)$,
otherwise the policy violates the given constraints.

From \cref{eqn:collision_selection} (our stoppable set selection strategy) and \cref{eqn:allstoppable_sets_includedby_KN},
all the possible states $\dot s^2_l \textrm{ s.t. } i < l \leq N$ can be categorized into one of the following two sets:
\begin{itemize}
  \item $\mathcal K_{j_{stop}, l}$ , that are `safe' (guaranteed as no collision), and
  \item ${\mathcal K}_{N, l}  \cap {\overline {\mathcal K}_{{j_{stop}}, l}} \triangleq {\mathcal K_{unsafe, l}}$, that are not `safe' (there exists a human motion that can collide with the robot).
\end{itemize}

At the forward step of the execution phase at the stage $i$, there are two cases where:
\begin{enumerate}
    \item $\mathcal K_{j_{stop}, i+1}$ becomes a non-active constraint.
    \item $\mathcal K_{j_{stop}, i+1}$ becomes an active constraint.
\end{enumerate}
In the case 1: TOPP-RA's forward pass generates the most greedy control input $u^*_i \in \mathcal U_i(x)$ via maximization of $u$, which contradicts with the assumption that there exists the strictly faster control $u^{**}_i$ than $u^*_i$ (see \cref{fig:proof2a} for illustration).

In the case 2: $u^{**}_i$ drives the robot state $x_{i+1}$ at the stage $(i+1)$ out of $\mathcal K_{j_{stop}, i+1}$ with $x_{i+1} = x_i + 2\triangle_i u^{**}_i$, which is in ${\mathcal K_{unsafe, i+1}}$.
This is unsafe and contradicts the initial assumption that $u^{**}_i$ is safe (see \cref{fig:proof2b} for illustration). $\square$
\end{appendices}

\bibliographystyle{unsrt}
\bibliography{sample}

\end{document}